\pdfoutput=1

\documentclass[11pt]{article}

\usepackage[]{emnlp2024_submission_arxiv}

\usepackage{times}
\usepackage{latexsym}
\usepackage[export]{adjustbox}
\usepackage{xcolor}

\newcommand{\tbu}[1]{\textbf{#1}} 

\usepackage[T1]{fontenc}

\usepackage[utf8]{inputenc}

\usepackage{microtype}

\usepackage{inconsolata}

\usepackage{multirow}
\usepackage{booktabs}

\usepackage{enumitem}

\usepackage{tikz}
\newcommand*\circled[1]{\tikz[baseline=(char.base)]{
            \node[shape=circle,draw,inner sep=1pt] (char) {#1};}}

\title{NLIP\_Lab-IITH Low-Resource MT System for WMT24 Indic MT Shared Task}

\author{Pramit Sahoo \hspace{1em} Maharaj Brahma \hspace{1em} Maunendra Sankar Desarkar \\
       Natural Language and Information Processing Lab 
       (NLIP)\\ Indian Institute of Technology Hyderabad \\ Hyderabad, India\\  \texttt{\{ai23mtech14004, cs23resch01004\}@iith.ac.in}, \texttt{maunendra@cse.iith.ac.in}}

\begin{document}
\maketitle

\begin{abstract}
In this paper, we describe our system for the WMT 24 shared task of  Low-Resource Indic Language Translation. We consider eng $\leftrightarrow$ \{as, kha, lus, mni\} as participating language pairs. In this shared task, we explore the finetuning of a pre-trained model motivated by the pre-trained objective of aligning embeddings closer by alignment augmentation \cite{lin-etal-2020-pre} for 22 scheduled Indian languages. Our primary system\footnote{Our code, models, and generated translations are available here: \url{https://github.com/pramitsahoo/WMT2024-LRILT}} is based on language-specific finetuning on a pre-trained model. We achieve  chrF2 scores of 50.6, 42.3, 54.9, and 66.3 on the official public test set for eng$\rightarrow$as,  eng$\rightarrow$kha, eng$\rightarrow$lus, eng$\rightarrow$mni respectively. We also explore multilingual training with/without language grouping and layer-freezing.
\end{abstract}

\section{Introduction}
The ``Shared Task: Low-Resource Indic Language Translation'' for WMT 2024 \cite{pakray2024findings} extends the efforts initiated in WMT 2023 \cite{pal-etal-2023-findings}, which garnered significant participation from the global community. Recent advancements in machine translation (MT), particularly through techniques like multilingual training and transfer learning, have expanded the scope of MT systems beyond high-resource languages \cite{johnson-etal-2017-googles}. However, low-resource languages continue to present substantial challenges due to the scarcity of parallel data required for effective training \cite{siddhant-etal-2020-leveraging, Wang2022UnderstandingAM}. The shared task focuses on low-resource Indic languages with limited data from diverse language families: Assamese (as), Mizo (lus), Khasi (kha), and Manipuri (mni). The task aims to improve translation quality for the English$\Leftrightarrow$Assamese, English$\Leftrightarrow$Mizo, English$\Leftrightarrow$Khasi, and English$\Leftrightarrow$Manipuri given the data provided in the constrained setting.

To address the challenges inherent in translating low-resource languages, participants are encouraged to explore several strategies. First, leveraging monolingual data is essential for enhancing translation quality, especially in the absence of sufficient parallel data. Second, multilingual approaches offer the potential for cross-lingual transfer, where knowledge from high-resource languages can be applied to low-resource pairs \cite{sen-etal-2019-multilingual}. Third, transfer learning provides a mechanism for adapting pre-trained models from high-resource languages to low-resource settings \cite{wang-etal-2020-balancing}. Lastly, innovative techniques tailored to low-resource scenarios, such as data augmentation and language-specific fine-tuning, are crucial for improving performance.

In this paper, we describe our system for the WMT 2024 shared task, focusing on fine-tuning two pre-trained models: IndicRASP and IndicRASP Seed\footnote{These pre-trained models were developed for WAT 2024 MultiIndicMT shared task by the authors.}. IndicRASP model is pre-trained with the objective of aligning embeddings inspired by alignment augmentation \cite{lin-etal-2020-pre} on 22 Indic languages. Our primary approach involves language-specific fine-tuning, leveraging multilingual training setups, language grouping, and layer freezing. We set up experiments in both bilingual and multilingual settings. We achieve BLEU scores of 20.1 for English$\rightarrow$Assamese, 19.1 for English$\rightarrow$Khasi, 30.0 for English$\rightarrow$Mizo, and 35.6 for English$\rightarrow$Manipuri on the public test set, demonstrating the effectiveness of our approach. Specifically, language-specific fine-tuning yielded significant improvements in translation quality, while multilingual setups provided balanced performance across all language pairs. Language grouping and layer freezing are effective techniques for preserving pre-trained knowledge and mitigating the challenges of multilinguality. Our results highlight the importance of tailored fine-tuning strategies for low-resource languages and show the potential of using alignment-augmented pre-trained models to improve translation quality in low-resource settings.

\section{Data}
In this section, we present the details of the IndicNECorp1.0 dataset provided by the IndicMT shared task\footnote{\url{https://www2.statmt.org/wmt24/indic-mt-task.html}} organizers.

\begin{table}[!htp]
\centering
\scriptsize
\begin{tabular}{c|c|c|c}
\toprule
\textbf{Language pair} & \textbf{Script} & \textbf{Dataset} & \textbf{\#parallel sents}
\\ \midrule
\multirow{3}{*}{English-Assamese} & \multirow{3}{*}{Bengali} & Training  & 50000  \\ \cline{3-4} 
 &  & Validation    &  2000  \\ \cline{3-4} 
 &  & Test  &  2000 \\ \midrule
\multirow{3}{*}{English-Khasi} & \multirow{3}{*}{Latin} & Training   & 24000  \\ \cline{3-4} 
 &  & Validation  & 1000   \\ \cline{3-4} 
 &  & Test  & 1000 \\ \midrule
\multirow{3}{*}{English-Manipuri} & \multirow{3}{*}{Bengali} & Training   & 21687  \\ \cline{3-4} 
 &  & Validation   & 1000 \\ \cline{3-4} 
 &  & Test  & 1000 \\ \midrule
\multirow{3}{*}{English-Mizo} & \multirow{3}{*}{Latin} & Training   & 50000  \\ \cline{3-4} 
 &  & Validation  & 1500  \\ \cline{3-4} 
 &  & Test  &  2000 \\ \bottomrule
\end{tabular}
\caption{Parallel dataset details. Script refers to the script of the Indic language.}
\label{tab:dataset_details}
\end{table}

\subsection{Monolingual Data}
The official data also includes monolingual data for four languages. The dataset comprises approximately 2.6M sentences for Assamese, 0.1M for Khasi, 2M for Mizo, and 1M for Manipuri.

\subsection{Parallel Data}
The dataset includes four bilingual pairs between English and Indic languages\footnote{Language code as per the dataset provided}: English (en) - Assamese (as), English (en) - Khasi (kha), English (en) - Mizo (lus), and English (en) - Manipuri (mni). These languages are mainly spoken in the North-eastern part of India. The English-Assamese and English-Mizo training sets contain 50k parallel sentences each, while the English-Khasi and English-Manipuri training sets contain 24k and 21.6k parallel sentences, respectively. Dataset statistics are presented in Table \ref{tab:dataset_details}.

\section{Approach}
In this section, we briefly describe our approaches. We explore transfer learning, language grouping, and layer-freezing techniques.

\subsection{Transfer Learning}
We explore transfer learning based on two pre-trained models ``IndicRASP" and ``IndicRASP Seed," which is a fine-tuned model of IndicRASP on small and high-quality data. Particularly, the pre-trained model is trained on agreement-based objective \cite{lin-etal-2020-pre, yang-etal-2020-csp} for Indic languages. The model is pre-trained in 22 scheduled Indic languages using a subset of the Bharat Parallel Corpus Collection (BPCC) dataset \cite{gala2023indictrans}. Out of these 22 languages, two of the shared task languages, Assamese and Manipuri, are part of the pre-training. Alignment augmentation is performed using bi-lingual dictionaries from MUSE\footnote{\url{https://github.com/facebookresearch/MUSE\#ground-truth-bilingual-dictionaries}} \cite{conneau2017word} and GATITOS\footnote{\url{https://github.com/google-research/url-nlp/tree/main/gatitos}}.



\subsection{Language Grouping}
We explore the effect of grouping languages based on script similarity in a multilingual setup. Although our primary focus is on bilingual models, for language grouping experiments, we utilize a multilingual approach where languages sharing similar scripts are trained together. This approach is motivated by the idea that joint training with similar languages can improve translation quality due to shared vocabulary and linguistic properties \citep{jiao-etal-2022-tencents, gala2023indictrans}.

\begin{itemize}[leftmargin=*]
    \item \textbf{Group 1} (Bengali script): Assamese and Manipuri
    \item \textbf{Group 2} (Latin script): Khasi and Mizo
\end{itemize}

\subsection{Layer Freezing}

We explored layer-freezing approaches for IndicTrans2 Distilled and IndicRASP Seed models.

\noindent\textbf{Frozen Encoder.}
In this approach, we freeze the encoder components during the fine-tuning process to preserve their pre-trained weights from the parent model while the embedding and decoder components are updated.


\noindent\textbf{Frozen Embedding + Encoder.} In this setup, we keep the embedding and encoder frozen during fine-tuning to preserve their pre-trained weights while updating only the parameters of the rest of the layers.

\begin{table*}[!htp]
\centering
\scriptsize
\scalebox{0.85}{
\begin{tabular}{l| cccc | cccc }
    \toprule
\multirow{2}{*}{\textbf{Models}} 
        & \multicolumn{4}{c|}{\textbf{ English $\rightarrow$ Indic}} & \multicolumn{4}{c}{\textbf{Indic $\rightarrow$ English}}   \\ 
        & \textbf{as}  & \textbf{kha}  & \textbf{lus}  & \textbf{mni} & \textbf{as} & \textbf{kha} & \textbf{lus} & \textbf{mni}    \\
\midrule   
    \multicolumn{9}{c}{\textbf{\textsc{Bilingual Setup}}} \\
    \midrule 
    \textsc{IndicTrans2 Distilled FT on Bilingual data} & 49.5 & 24.9 & 29.1 & 60.1	 & 50.9 & 21.1 & 22.0 & 61.9 \\ 
    
    \midrule 
    \textsc{IndicRASP FT on Bilingual data} & 49.9 & 42.2 & 36.5 & 65.8 & 50.1 & 36.1 & 49.4 & 67.7  \\ 
    \midrule 
    \textsc{IndicRASP Seed FT on Bilingual data} & 50.6 & 42.3 & 54.9 & 66.3 & 52.8 & 36.1 & 25.1 & 67.9  \\ 
    \midrule


\multicolumn{9}{c}{\textbf{\textsc{Multilingual Setup}}} \\
    \midrule 
    \textsc{IndicRASP FT on Multilingual data} & 49.8 & 34.6 & 51.5 & 63.2 & 51.2 & 36.0 & 46.5 & 65.3  \\
    \midrule 
    \textsc{IndicRASP Seed FT on Multilingual data} & 48.7 & 34.6 & 50.2 & 62.2 & 52.2 & 35.3 & 44.3 & 65.1  \\
    \midrule


\multicolumn{9}{c}{\textbf{\textsc{Multilingual Model FT on Bilingual Data}}} \\
    \midrule 
    \textsc{IndicRASP Multilingual Model FT on Bilingual data} & 49.3 & 42.4 & 54.7 & 65.8 & 50.9 & 36.3 & 46.8 & 67.4 \\ 
    \midrule

    \multicolumn{9}{c}{\textbf{\textsc{Layer Freezing}}} \\
    \midrule
    \textsc{IndicTrans2 Distilled FT with Frozen Encoder}  & 47.4 & 24.4 & 28.0 & 57.8 & 48.7 & 19.8 & 18.7 & 58.8 \\ 
    \midrule
    \textsc{IndicRASP Seed FT with Frozen Encoder} & 50.4 & 41.3 & 48.6 & 63.4 & 52.6 & 26.4 & 34.2 & 65.3  \\ 

    \midrule
    \textsc{IndicTrans2 Distilled FT with Frozen Embedding \& Encoder } & 46.7 & 23.1 & 9.2 & 15.9 & 48.8 & 20.2 & 19.6 & 58.1 \\ 
    \midrule
    \textsc{IndicRASP Seed FT with Frozen Embedding \& Encoder} & 50.5 & 41.2 & 45.8 & 62.4 & 52.9 & 25.9 & 29.6 & 64.1  \\ 
    \midrule
    
    \multicolumn{9}{c}{\textbf{\textsc{Language Grouping}}} \\
    \midrule
    \textsc{IndicRASP FT with Script Similarity} & 50.2 & 35.0 & 52.1 & 63.3 & 52.6 & 36.4 & 46.5 & 66.0 \\ 

    \midrule
    \textsc{IndicRASP Seed Model FT with Script Similarity} & 50.3 & 34.9 & 53.5 & 63.6 & 53.6 & 36.8 & 47.4 & 66.8 \\

\bottomrule
\end{tabular}}
\vspace*{-0.2cm} 
\caption{\small chrF2 scores  on IndicMT WMT24 shared task public test set.}
\label{tab:resultschrF2}
\end{table*}

\begin{table*}[!h]
\centering
\scriptsize
\scalebox{0.85}{
\begin{tabular}{l| cccc | cccc}
    \toprule
\multirow{2}{*}{\textbf{Models}} 
        & \multicolumn{4}{c|}{\textbf{ English $\rightarrow$ Indic}} & \multicolumn{4}{c}{\textbf{Indic $\rightarrow$ English}}   \\ 
        & \textbf{as}  & \textbf{kha}  & \textbf{lus}  & \textbf{mni} & \textbf{as} & \textbf{kha} & \textbf{lus} & \textbf{mni}       \\
\midrule   
    \multicolumn{9}{c}{\textbf{\textsc{Bilingual Setup}}} \\
    \midrule 
    \textsc{IndicTrans2 Distilled FT on Bilingual data} & 18.0 & 9.3 & 13.6 & 21.6 & 26.3 & 2.7 & 5.0 & 36.2  \\ 
    
    \midrule 
    \textsc{IndicRASP FT on Bilingual data} & 20.5 & 18.9 & 13.1 & 33.9 & 20.0 & 14.4 & 29.1 & 43.6  \\ 
    \midrule 
    \textsc{IndicRASP Seed FT on Bilingual data} & 20.1 & 19.1 & 30.0 & 35.6 & 27.4 & 14.1 & 6.0 & 44.1  \\ 
    \midrule


\multicolumn{9}{c}{\textbf{\textsc{Multilingual Setup}}} \\
    \midrule 
    \textsc{IndicRASP FT on Multilingual data} & 18.7 & 13.5 & 25.8 & 29.0 & 25.8 & 14.1 & 25.4 & 39.3  \\ 
    \midrule 
    \textsc{IndicRASP Seed FT on Multilingual data} & 17.1 & 13.2 & 24.4 & 27.2 & 26.7 & 14.1 & 23.3 & 38.3  \\
    \midrule


\multicolumn{9}{c}{\textbf{\textsc{Multilingual Model FT on Bilingual Data}}} \\
    \midrule 
    \textsc{IndicRASP Multilingual Model FT on Bilingual data} & 19.1 & 19.0 & 29.7 & 34.7 & 25.8 & 14.8 & 26.1 & 43.5  \\ 
    \midrule

    \multicolumn{9}{c}{\textbf{\textsc{Layer Freezing}}} \\
    \midrule
    \textsc{IndicTrans2 Distilled FT with Frozen Encoder} & 15.6 & 8.9 & 13.1 & 19.6 & 22.7 & 1.5 & 3.0 & 31.3  \\ 
    \midrule
    \textsc{IndicRASP Seed FT with Frozen Encoder} & 19.7 & 18.1 & 22.4 & 29.0 & 26.8 & 5.6 & 15.2 & 40.7  \\ 

    \midrule
    \textsc{IndicTrans2 Distilled FT with Frozen Embedding \& Encoder} & 14.8 & 8.3 & 2.6 & 1.3 & 22.7 & 1.9 & 3.8 & 30.5  \\ 
    \midrule
    \textsc{IndicRASP Seed FT with Frozen Embedding \& Encoder} & 19.4 & 17.7 & 19.7 & 27.2 & 26.9 & 5.4 & 10.9 & 37.9  \\ 
    \midrule
    
    \multicolumn{9}{c}{\textbf{\textsc{Language Grouping}}} \\
    \midrule
    \textsc{IndicRASP FT with Script Similarity} & 19.1 & 13.8 & 26.6 & 28.9 & 26.9 & 14.6 & 25.5 & 39.8  \\ 

    \midrule
    \textsc{IndicRASP Seed Model FT with Script Similarity} & 19.4 & 14.1 & 28.6 & 29.4 & 28.3  &  14.8 & 26.4  & 40.6  \\

\bottomrule
\end{tabular}}
\vspace*{-0.2cm} 
\caption{\small BLEU scores on IndicMT WMT24 shared task public test set.}
\label{tab:resultsBLEU}
\end{table*}

\section{Experimental Setup}
\noindent\textbf{Settings.} We fine-tune pre-trained checkpoints: ``IndicRASP'' and ``IndicRASP Seed'' models on official parallel data using the Adam optimizer \cite{kingma2014adam} with $\beta_1$ set to 0.9 and $\beta_2$ set to 0.98. We set the initial warmup learning rate to 1e-07 and the learning rate to 3e-5, with a warmup step of 4000. We train the models with a dropout rate of 0.3 and a label smoothing rate of 0.1. All experiments are conducted on a single NVIDIA A100 GPU. We use a maximum token count of 512 per batch, accumulating gradients over two steps to simulate a larger batch size. The model is trained for up to 1,000,000 updates. We save checkpoints every 2500 updates. We employed a patience of 10 for early stopping.

\noindent\textbf{Evaluation Metrics.}
We use the official dev and test sets of IndicNECorp1.0 for validation and evaluation. We evaluate using BLEU \cite{papineni-etal-2002-bleu}, chrF \cite{popovic-2015-chrf}, and chrF++ \cite{popovic-2017-chrf} metrics. We use the SacreBLEU toolkit \cite{post-2018-call} to perform our evaluation\footnote{\raggedright SacreBLEU signature: \texttt{nrefs:1|case:mixed|eff:no|tok:13a
|smooth:exp|version:2.3.1}} with a chrF word order of 2. Additionally, as per the evaluation metrics used by the organizers, we report results on TER \cite{snover-etal-2006-study}, RIBES \cite{isozaki-etal-2010-automatic}, and COMET \cite{rei-etal-2022-comet} for our primary and contrastive submissions. 

\noindent\textbf{Models.} We conducted our experiments in both bilingual and multilingual settings. In the bilingual setup, we fine-tuned the IndicTrans2 Distilled model \cite{gala2023indictrans}, IndicRASP, and IndicRASP Seed models for both English to Indic and Indic to English directions. In the multilingual setup, we fine-tuned pre-trained checkpoints of IndicRASP and IndicRASP Seed for both directions. Inspired by \citet{chiang-etal-2022-breaking}, we initialized the bilingual model with a fine-tuned multilingual model for both English to Indic and Indic to English.

For experiments with layer freezing, we fine-tune pre-trained checkpoints of IndicTrans2 Distilled and IndicRASP Seed models. Particularly, we perform experiments by freezing the embeddings and encoder and only the encoder component for both English to Indic and Indic to English directions. We conduct all layer-freezing experiments in a bilingual setup. For language grouping experiments, we fine-tune the IndicRASP and IndicRASP Seed models based on script similarity in a multilingual setup.

\section{Results and Discussions}
In this section, we report our experimental results and describe our primary and contrastive submissions. The results for our primary and contrastive systems are shown in Table \ref{tab:submissionResults}. Tables \ref{tab:resultschrF2}, \ref{tab:resultsBLEU}, and \ref{tab:resultschrF++} reports the chrF2, BLEU, and chrF++ scores respectively.
\begin{table}[h]
\centering
\scriptsize
\scalebox{0.95}{
    \begin{tabular}{c c c c c c}
    \toprule
         &  \textbf{BLEU} & \textbf{chrF2} & \textbf{TER} & \textbf{RIBES} & \textbf{COMET} \\
         \midrule
         \multicolumn{6}{c}{\textbf{\textsc{Primary}}} \\
        \midrule
        en$\rightarrow$as &  20.1 & 50.6  & 66.0 & 0.5543 & 0.8090 \\
        en$\rightarrow$kha & 19.1  & 42.3 & 63.5 & 0.6470 & 0.6817 \\
        en$\rightarrow$lus & 30.0  & 54.9 & 50.0 & 0.6764 & 0.7105 \\
        en$\rightarrow$mni & 35.6  & 66.3 & 50.5 & 0.6995 & 0.7669 \\
        \midrule
        as$\rightarrow$en & 27.4  & 52.8  & 65.3 & 0.6749 & 0.7854 \\
        kha$\rightarrow$en & 14.4 & 36.1 & 82.0 & 0.5601 & 0.5773 \\
        lus$\rightarrow$en & 29.1 & 49.4 & 66.7 & 0.6436 & 0.7004 \\
        mni$\rightarrow$en & 44.1  & 67.9 & 50.2 & 0.7894 & 0.8162 \\
        \midrule
        \multicolumn{6}{c}{\textbf{\textsc{Contrastive}}} \\
        \midrule
        en$\rightarrow$as & 20.5  & 49.9  & 67.2 & 0.5356 & 0.8043 \\
        en$\rightarrow$kha & 18.9  & 42.2  & 63.5 & 0.6499 & 0.6791\\
        en$\rightarrow$lus &  13.1  & 36.5  & 73.8 & 0.4357 & 0.6462 \\
        en$\rightarrow$mni &  33.9  & 65.8 & 50.5 & 0.6972 & 0.7672 \\
        \midrule
        as$\rightarrow$en & 25.8  & 51.2  & 66.8 & 0.6744 & 0.7802 \\
        lus$\rightarrow$en & 25.4  & 46.5  & 69.0 & 0.6307 & 0.6882\\
        mni$\rightarrow$en & 39.3  & 65.3  & 52.4 & 0.7806 & 0.8034\\
        \bottomrule
    \end{tabular}
    }
    \caption{\small Submission results on the IndicMT WMT24 public test set.}
    \label{tab:submissionResults}
\end{table}
\begin{enumerate}[label=\protect\circled{\arabic*}]
\item \textbf{English $\rightarrow$ Indic:} Our primary English to Indic systems are language pair-specific (bilingual models) fine-tuned on pre-trained IndicRASP Seed, achieving chrF2 scores of 50.6, 42.3, 54.9, and 66.3 for Assamese, Khasi, Mizo, and Manipuri respectively. For the contrastive systems, we consider a bilingual model fine-tuned on a pre-trained IndicRASP checkpoint. The contrastive system achieves chrF2 scores of 49.9, 42.2, 36.5, and 65.8 for Assamese, Khasi, Mizo, and Manipuri, respectively. The detailed primary and contrastive system results are reported in Table \ref{tab:submissionResults}.

\item \textbf{Indic $\rightarrow$ English:} Our primary Indic-to-English systems for Assamese and Manipuri are bilingual models fine-tuned on the pre-trained IndicRASP Seed model, each achieving chrF2 scores of 52.8 and 67.9, respectively. Similarly, for Khasi and Mizo, our primary systems are bilingual models fine-tuned on a pre-trained IndicRASP checkpoint, achieving a chrF2 score of 36.1 and 49.4, respectively.

For the contrastive Indic-to-English system, we submit a multilingual system fine-tuned on the pre-trained checkpoint of the IndicRASP model, achieving chrF2 scores of 51.2, 36.0, 46.5, and 65.3 for Assamese, Khasi, Mizo, and Manipuri respectively. Table \ref{tab:submissionResults} shows the detailed scores in various metrics.
\end{enumerate}

\begin{table*}[!h]
\centering
\scriptsize
\scalebox{0.9}{
\begin{tabular}{l| cccc | cccc}
    \toprule
\multirow{2}{*}{\textbf{Models}} 
        & \multicolumn{4}{c|}{\textbf{ English $\rightarrow$ Indic}} & \multicolumn{4}{c}{\textbf{Indic $\rightarrow$ English}}  \\
        & \textbf{as}  & \textbf{kha}  & \textbf{lus}  & \textbf{mni} & \textbf{as} & \textbf{kha} & \textbf{lus} & \textbf{mni}         \\
\midrule   
    \multicolumn{9}{c}{\textbf{\textsc{Bilingual Setup}}} \\
    \midrule 
    \textsc{IndicTrans2 Distilled FT on Bilingual data} & 45.8 & 25.6 & 30.3 & 55.4 & 49 & 20 & 21 & 59.6  \\ 
    
    \midrule 
    \textsc{IndicRASP FT on Bilingual data} & 46.4 & 41.3 & 35.2 & 61.8 & 46.5 & 35.3 & 48.2 & 65.4  \\
    \midrule 
    \textsc{IndicRASP Seed FT on Bilingual data} & 47 & 41.4 & 53.2 & 62.3 & 50.6 & 35.3 & 24 & 65.7  \\
    \midrule


\multicolumn{9}{c}{\textbf{\textsc{Multilingual Setup}}} \\
    \midrule 
    \textsc{IndicRASP FT on Multilingual data} & 46.2 & 33.4 & 49.8 & 58.9 & 49.1 & 35.2 & 45.4 & 63  \\
    \midrule 
    \textsc{IndicRASP Seed FT on Multilingual data} & 45.1 & 33.4 & 48.5 & 57.9 & 50.1 & 34.6 & 43.2 & 62.6  \\
    \midrule


\multicolumn{9}{c}{\textbf{\textsc{Multilingual Model FT on Bilingual Data}}} \\
    \midrule 
    \textsc{IndicRASP Multilingual Model FT on Bilingual data} & 45.7 & 41.5 & 53.1 & 61.9 & 48.8 & 35.5 & 45.7 & 65.2  \\
    \midrule

    \multicolumn{9}{c}{\textbf{\textsc{Layer Freezing}}} \\
    \midrule
    \textsc{IndicTrans2 Distilled FT with Frozen Encoder} & 43.7 & \tbu{25.1} & 29.3 & \tbu{53} & \tbu{46.7} & \tbu{18.5} & 17.6 & \tbu{59.8}  \\
    \midrule
    \textsc{IndicRASP Seed FT with Frozen Encoder} & 46.8 & \tbu{40.3} & 46.9 & \tbu{59.1} & \tbu{50.4} & \tbu{25.3} & 33.1 & \tbu{63}  \\

    \midrule
    \textsc{IndicTrans2 Distilled FT with Frozen Encoder \& Embeddings} & 43 & \tbu{24} & 11.3 & \tbu{13.1} & \tbu{46.8} & \tbu{18.9} & 18.5 & \tbu{55.6}  \\
    \midrule
    \textsc{IndicRASP Seed FT with Frozen Encoder \& Embeddings} & 46.8 & \tbu{40.2} & 44.1 & \tbu{58} & \tbu{50.6} & \tbu{24.9} & 28.6 & \tbu{61.7}  \\
    \midrule
    
    \multicolumn{9}{c}{\textbf{\textsc{Language Grouping}}} \\
    \midrule
    \textsc{IndicRASP FT with Script Similarity} & 46.6 & 33.8 & 50.4 & 59 & 50.4  &  35.6 & 45.4  & 63.6  \\

    \midrule
    \textsc{IndicRASP Seed Model FT with Script Similarity} & 46.7 & 33.7 & 51.8 & 59.4 & 51.5  & 36 & 46.3  & 64.4  \\
\bottomrule
\end{tabular}
}
\vspace*{-0.2cm} 
\caption{\small chrF2++ scores on IndicMT WMT24 shared task public test set.}
\label{tab:resultschrF++}
\end{table*}
\noindent\textbf{Bilingual vs. Multilingual.} We observe that fine-tuning the pre-trained IndicRASP Seed outperforms the IndicRASP model, likely due to the continued pre-training on a small, high-quality dataset in the IndicRASP Seed model. Bilingual models perform better than multilingual models, showing a +4.1 and +7.7 chrF2 score improvement for English to Manipuri and English to Khasi, respectively.

Bilingual models initialized with the weights from multilingual models show improvement over the standalone multilingual models, achieving a +7.8 chrF2 score for English to Khasi. This suggests that initializing bilingual models can be helpful in low-resource settings.

\noindent\textbf{Language Grouping.} We observe that script-based language grouping shows improvements over a standalone multilingual model with +1.6, +0.3, +3.3, and +1.4 for English to Assamese, Khasi, Mizo, and Manipuri, respectively. It suggests that grouping languages based on script similarity can be effective in addressing the curse of multilinguality.

\noindent\textbf{Layer Freezing.} We observe that freezing only the encoder yields better chrF2 scores compared to freezing both the embedding and the encoder. However, layer freezing underperforms compared to full parameter fine-tuned bilingual models.

\section{Conclusion}
In this paper, we describe NLIP Lab's Indic low-resource machine translation systems for the WMT24 shared task. We explore the translation capabilities of the alignment-augmented pre-trained
model, IndicRASP and IndicRASP Seed, to enhance translation quality for low-resource Indic languages. Experimentally, we found that the IndicRASP model performs better than the IndicTrans2 Distilled model. Additionally, we experiment with layer-freezing and language grouping techniques. In the future, we will focus on refining these techniques and utilizing monolingual data to enhance MT performance for low-resource Indic languages.

\section*{Limitations}
The pre-trained models use bilingual dictionaries whose domain might differ from the shared task training corpus. Additionally, the considered pre-trained models cover only a limited number of shared task languages. Our submission does not utilize the provided monolingual data, which could further improve model performance through backtranslation.

\section*{Acknowledgements}
We express our gratitude to the reviewer for providing us with valuable feedback and suggestions for improving the readability of the paper. We also thank the Department of Artificial Intelligence and Department of Computer Science and Engineering, Indian Institute of Technology Hyderabad, for providing the necessary computing resources to conduct the experiments.

\bibliography{emnlp2024_submission_arxiv}
\bibliographystyle{acl_natbib}

\appendix



\end{document}